\documentclass[conference]{IEEEtran}
\usepackage{amsthm}
\newtheorem{rec}{Requirement}
\IEEEoverridecommandlockouts
\usepackage{cite}
\usepackage{amsmath,amssymb,amsfonts}
\usepackage{algorithm}
\usepackage{algpseudocode}
\usepackage{graphicx}
\usepackage{textcomp}
\usepackage{xcolor}
\usepackage{subcaption}

\newtheorem{definition}{Definition}

\usepackage{tikz}
\usetikzlibrary{shapes.geometric, arrows.meta, positioning, calc}

\begin{document}

\title{Adaptive Coverage Algorithms for Mobile Ad Hoc Networks in Dynamic Worlds}

\author{\IEEEauthorblockN{Hugo Gandhi}
\IEEEauthorblockA{\textit{CentraleSupélec} \\
Gif-sur-Yvette, France \\
hugo.gandhi@student-cs.fr}
\and
\IEEEauthorblockN{Edwin Meriaux}
\IEEEauthorblockA{
\textit{CentraleSupélec}\\
Gif-sur-Yvette, France \\
edwin.meriaux@mail.mcgill.ca}
\and
\IEEEauthorblockN{Antonio Loría}
\IEEEauthorblockA{
\textit{CNRS}\\
Gif-sur-Yvette, France \\
antonio.loria@cnrs.fr}
}

\maketitle

\begin{abstract}

This paper extends an existing framework for the coordination of mobile agents in unknown worlds, by considering the possibility of dynamic instability. In the original problem, known as the \textit{Partially Observable Cooperative Guard Art Gallery Problem} (POCGAGP) \cite{meriaux2025manet}, the multi-agent system must establish a communication network which completely covers an initially unknown constrained space. However, the POCGAGP fails to account for the unpredictability often seen in real-world deployment. To address this, the dynamic case (DPOCGAGP) introduces two settings which can influence the mobile network during deployment: \textit{Agent Loss} and \textit{World Change}, which represent dynamic changes to the world. Within this study, we present advancements to two state-of-the-art algorithms, the Centralized Agent-Directed Exploration for Network Coverage (CADENCE), and its decentralized analogue, DADENCE, to provide solutions in these cases. We then compare their performance against another state-of-the-art benchmark algorithm \cite{howard2002ISDA}, and show that the adapted algorithms outperform across our benchmark test set, which comprises over 4500 test cases of varied size and structure.

\end{abstract}


\section{Introduction}

As multi-robot and drone systems become increasingly prevalent, ensuring robust communication across distributed agents is critical to enabling reliable coordination and task execution in real-world environments. Reliable and flexible connectivity is of particular interest for Mobile Ad Hoc Networks (MANETs), which consist of interconnected mobile nodes that operate without fixed infrastructure.

The dynamic topology and self-organizing nature of MANETs have led to the development of several different routing and deployment protocols \cite{sharma2019study}, which can be used to obtain complete coverage of a sensor field \cite{wang2011coverage}. These protocols are becoming increasingly viable for use in unknown, spatially bounded environments; for example, in underground structures, such as tunnels or mines \cite{zhang2023uav}, or in post-disaster scenarios such as earthquake-damaged buildings, when  local communication infrastructure is damaged and communication is required for search-and-rescue
operations \cite{salamanca_survey}.

Meriaux et al. \cite{meriaux2025manet} conceptualize the challenge of deploying mobile sensor nodes in unknown worlds to create coverage networks, casting the problem into a computational-geometric framework known formally as the \textit{Partially Observable Cooperative Guard Art Gallery Problem} (POCGAGP).

\begin{figure}[h!]
    \centering
    \includegraphics[width=0.48\textwidth]{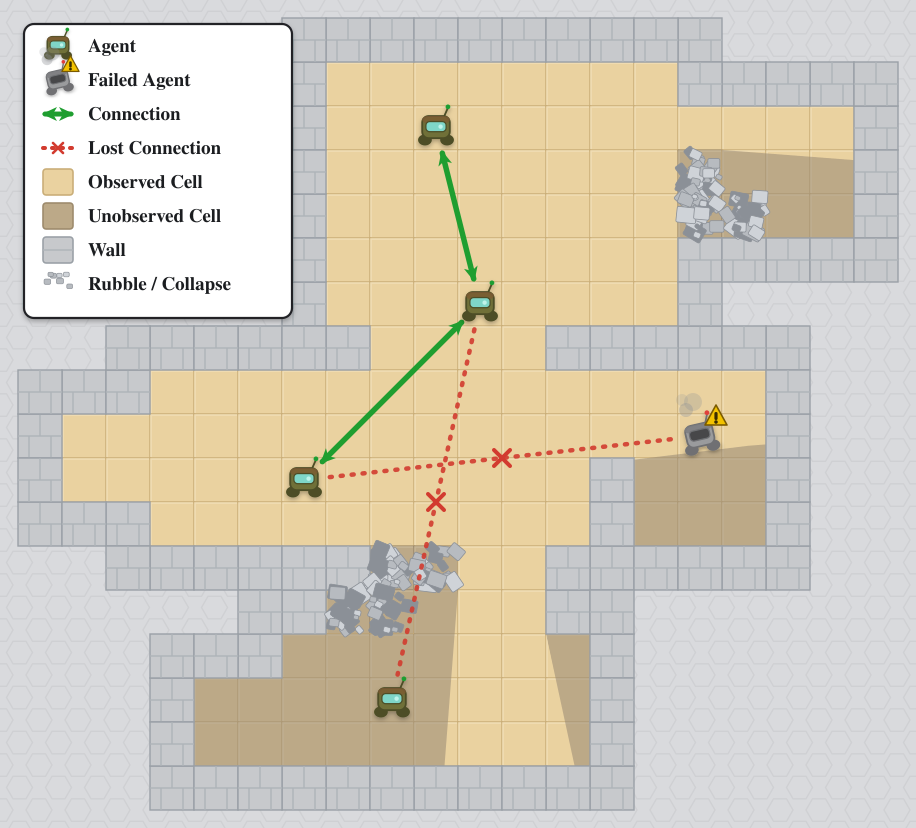}
    \caption{Robots attempting to form a MANET in a partially observed indoor environment with collapsed wall sections and a failed agent.}
    \label{fig:manet_pretty}
\end{figure}

This POCGAGP extends the \textit{Cooperative Guard Art Gallery Problem} (CGAGP), which
itself extends the classic \textit{Art Gallery Problem} (AGP). The AGP, proposed by Klee
and later solved by Chvátal \cite{honsberger1976mathematical}, asks how to place the
fewest observers in a polygon so that every point inside is visible. The CGAGP extends the
AGP to a cooperative system, in which the agents must not only cover the entire world but
also form a connected graph, i.e., every agent has a direct or indirect communication path
to all others. This lends itself well to the communication challenges inherent in multi-agent ad hoc networks, and directly motivates the formulation of the POCGAGP, in which a MANET must simultaneously explore, cover, and establish connectivity within an initially unknown,
constrained indoor space. To ground the problem in practical constraints, solutions to the
POCGAGP aim to minimize both the number of agents deployed and the total
deployment time, reflecting real-world hardware, cost, and energy limitations.

Two novel MANET algorithms are proposed in \cite{meriaux2025manet} to solve the POCGAGP:
a centralized approach, CADENCE, which relies on a single entity with global knowledge to
dictate agent actions, and a decentralized approach, DADENCE, which enables agents to make
decisions using only local information and communication.

While these algorithms display excellent performance in static simulations, the deployment of MANETs in real-world environments is frequently complicated by unpredictable disturbances not considered in the original problem formulation. In practical scenarios, robotic hardware failures, such as sudden power loss or mechanical faults, can abruptly sever existing communication links. Furthermore, dynamic obstacles like shifting rubble in disaster zones, moving crowds, or structural collapses can fundamentally alter the environment's topology, creating line-of-sight obstructions that degrade wireless signals \cite{aileen2021wifi} and prevent connected coverage.
An example of such a scenario is given in Figure \ref{fig:manet_pretty}, where structural collapse and unexpected agent failure partition the network and halt ongoing exploration. Considering these adverse conditions is critical to validating the robustness of MANET deployment algorithms in real-world environments. In this context, robustness is defined as the multi-agent system's capacity to dynamically adjust its connectivity logic, regain lost coverage, and maintain exploration while optimizing the use of available agents across potentially isolated subnetworks.

To address this gap, this study introduces the Dynamic POCGAGP (DPOCGAGP) that explicitly incorporates unpredictable environmental disturbances; namely, the arbitrary loss of agents from the network and sudden structural changes in the environment's topology. Subsequently, both CADENCE and DADENCE are modified to enable guaranteed recovery and resilience within these adversarial settings, and their performance is simulated and evaluated.

\section{Mathematical Background and the POCGAGP}
To formally present the DPOCGAGP, we first establish the necessary background.
The POCGAGP concerns the design of an online, multi-agent policy that progressively
discovers an initially unknown world $W \subset \mathbb{R}^N$.
In the network, each agent observes its own visible region (e.g., the area sensed by an ideal $360^\circ$ LiDAR) and shares these measurements with the others it can see.
Each agent must terminate in
a position connected to the greater network, such that the network collectively observes
all of $W$, subject to the following requirements.

\begin{rec}[Time Bound $T_{max}$] \label{req_time}
Agent deployment and movement must be completed within a finite time horizon of length $T_{max}$.
\end{rec}
\begin{rec}[Agent Bound $N_{max}$] \label{req_n}
The number of agents that may exist in the world at any time is bounded by a given
constant $N_{max}$.
\end{rec}
\begin{rec}[Consistent Connectivity] \label{req_con}
At any time, every action taken by the agents must maintain connectivity with the full network.
\end{rec}
\begin{rec}[Monotonic Growth of Coverage] \label{req_cov}
The actions taken by the MANET may only increase the collective field of view ($F_W$) over time.
\end{rec}

We further designate a fixed \textit{deployment point} $x_d \in W$, which represents the
network's connection to the outside world and the point through which agents enter or
leave. The vision from $x_d$ is assumed to contribute to the MANET's total vision, so the
collective vision is written $F_W(A \cup \{x_d\})$, where $A \subseteq W$ is the set of deployed agent positions.

\subsection{CADENCE}
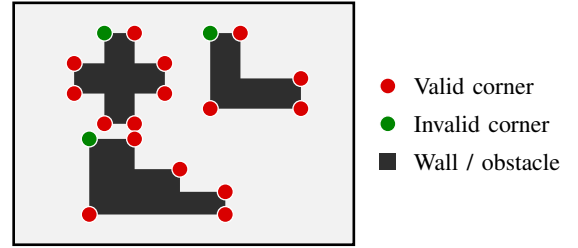
\begin{figure}[h!]
    \centering
    \begin{tikzpicture}[
        wall/.style={fill=black!82},
        valid/.style={circle, fill=red!85!black, draw=white, line width=0.5pt,
                      inner sep=0pt, minimum size=6pt},
        invalid/.style={circle, fill=green!52!black, draw=white, line width=0.5pt,
                        inner sep=0pt, minimum size=6pt},
    ]
        \fill[black!5] (0,0) rectangle (4.5,3.2);

        \fill[wall] (1.2,2.8)--(1.6,2.8)--(1.6,2.4)--(2.0,2.4)--(2.0,2.0)--(1.6,2.0)
                    --(1.6,1.6)--(1.2,1.6)--(1.2,2.0)--(0.8,2.0)--(0.8,2.4)--(1.2,2.4)--cycle;
        \fill[wall] (2.6,2.8)--(3.0,2.8)--(3.0,2.2)--(3.8,2.2)--(3.8,1.8)--(2.6,1.8)--cycle;
        \fill[wall] (1.0,1.4)--(1.6,1.4)--(1.6,1.0)--(2.2,1.0)--(2.2,0.7)--(2.8,0.7)
                    --(2.8,0.4)--(1.0,0.4)--cycle;

        \draw[line width=1pt] (0,0) rectangle (4.5,3.2);

        \foreach \p in {
            (1.6,2.8),(2.0,2.4),(2.0,2.0),(1.6,1.6),(1.2,1.6),(0.8,2.0),(0.8,2.4),
            (3.0,2.8),(3.8,2.2),(3.8,1.8),(2.6,1.8),
            (1.6,1.4),(2.2,1.0),(2.8,0.7),(2.8,0.4),(1.0,0.4)}
            \node[valid] at \p {};

        \foreach \p in {(1.2,2.8),(2.6,2.8),(1.0,1.4)}
            \node[invalid] at \p {};

        \begin{scope}[shift={(4.95,1.6)}]
            \node[valid]   at (0,0.5) {};
            \node[anchor=west, font=\small] at (0.22,0.5) {Valid corner};
            \node[invalid] at (0,0) {};
            \node[anchor=west, font=\small] at (0.22,0) {Invalid corner};
            \fill[wall] (-0.11,-0.61) rectangle (0.11,-0.39);
            \node[anchor=west, font=\small] at (0.22,-0.5) {Wall / obstacle};
        \end{scope}
    \end{tikzpicture}
    \caption{Illustration of a polygon $W$ with all invalid reflexive corners
    in green and valid corners in red.}
    \label{fig:all_270}
\end{figure}

CADENCE~\cite{meriaux2025manet} is a centralized algorithm that uses the collective observations gathered by all deployed agents to systematically compute the next optimal deployment targets, ensuring highly coordinated exploration.
This algorithm solves the POCGAGP by leveraging geometric heuristics. Specifically, it identifies ``valid corners", reflexive vertices of $270^\circ$ in an orthogonal map excluding the top-left corner of holes, as optimal locations for agents (see Figure \ref{fig:all_270} for an example). By iteratively deploying agents to these corners, CADENCE guarantees full coverage and connectivity.

\subsection{DADENCE}
Unlike CADENCE, which relies on a global planner, DADENCE~\cite{meriaux2025manet} allows agents to operate autonomously using only local information and communication with their peers. By sharing observations and following identical greedy rules, the decentralized agents compute actions independently to reach a consensus.  Specifically, the agents collectively identify a target point $x^*$ located on the border $B(F_W (A \cup \{x_d\}))$ (i.e., the edge of coverage, also denoted $\mathbf{B}$) of their current discrete field of view. $x^*$ is selected to minimize the sum of the shortest-path distances to all currently deployed agents and the deployment point.  Before executing any intended movement, each agent verifies that its action will not result in collisions, a reduction of the covered area, or the disconnection of the discrete visibility graph $G(A\cup\{x_d\})$. If an intended action violates these safety conditions, the agent reverts its action and remains in place until the conflict is resolved (e.g., through finding another valid action or spawning a new agent at $x_d$).

\subsection{Greedy Deallocation}
\label{subsec:deallocation}
To optimize the agent usage, both CADENCE and DADENCE can use a \textit{Greedy Deallocation} policy. As some agents may become effectively redundant as exploration progresses, a terminal agent is deemed ``unnecessary'' if its removal:
\begin{enumerate}
    \item Does not reduce the collective coverage of the world;
    \item Does not disconnect the visibility graph of the remaining agents.
\end{enumerate}
By running this deallocation step, CADENCE and DADENCE safely prune redundant nodes to minimize resource utilization while preserving the problem's mathematical constraints.
\subsection{Original Findings}
In their initial evaluation in \cite{meriaux2025manet}, both CADENCE and DADENCE vastly outperformed standard baseline methods for MANET deployment. A clear operational trade-off emerged between the two approaches: CADENCE achieves full coverage in significantly fewer steps by moving agents along direct paths to known goals, whereas DADENCE maximizes overall agent efficiency by deploying fewer nodes.
However, it is worth noting that a theoretical increase in resilience typically associated with a decentralized algorithm like DADENCE had not yet been explicitly proven in adversarial or dynamic settings. This limitation directly motivates the introduction of the DPOCGAGP framework.
\section{The DPOCGAGP}

This study introduces a \textbf{Dynamic} extension of the POCGAGP, denoted DPOCGAGP,
in which the environment and agent network evolve under unpredictable disturbances. The DPOCGAGP adds one requirement
to the four already defined:

\begin{rec}[Recovery] \label{robust}
During deployment, the system must autonomously recover from unpredictable
environmental disturbances, specifically:
    \begin{itemize}
        \item \textbf{Agent Loss} (Definition~\ref{def:agent_loss}): The sudden, permanent removal of a subset of agents
              from the network.
        \item \textbf{World Change} (Definition~\ref{def:world_change}): A structural change in the topology of the
              environment.
    \end{itemize}
\end{rec}

\begin{definition}[DPOCGAGP]\label{def:dpocgagp}
The DPOCGAGP is the POCGAGP extended with Requirement~\ref{robust}: at each step $t$,
the navigable world $W_t$ and agent set $A_t$ may change via an Agent Loss or World
Mutation event, and the policy must restore full connected coverage
$F_{W_t}(A_t \cup \{x_d\}) = W_t$ of the current world after each event.
\end{definition}

During recovery, solutions to the DPOCGAGP can be evaluated on their ability to adapt the
remaining agent network while minimizing both the time required to regain lost coverage and
connectivity, and the number of agents used. Furthermore, w.r.t.\ Requirement~\ref{req_n},
it is assumed that if an agent is destroyed, a replacement agent can be deployed from the
deployment point without incurring an additional cost against the network's maximum agent
budget ($N_{max}$). This is done to maintain enough agents such that a solution could exist.

\begin{definition}[Agent Loss]\label{def:agent_loss}
An \textit{Agent Loss} is defined as an instantaneous, permanent failure of one
or more agents in the network. Let $A_t$ represent the set of deployed agents at
a given discrete time step $t$. An Agent Loss that transitions the network from
$A_t$ to $A_{t+1}$ can be described by the removal of a randomly selected subset
of failed agents $A_{t,fail} \subseteq A_t$:
$$A_{t+1} = A_t \setminus A_{t,fail}$$
\end{definition}

\begin{figure}[h!]
    \centering
    \includegraphics[width=0.4\textwidth]{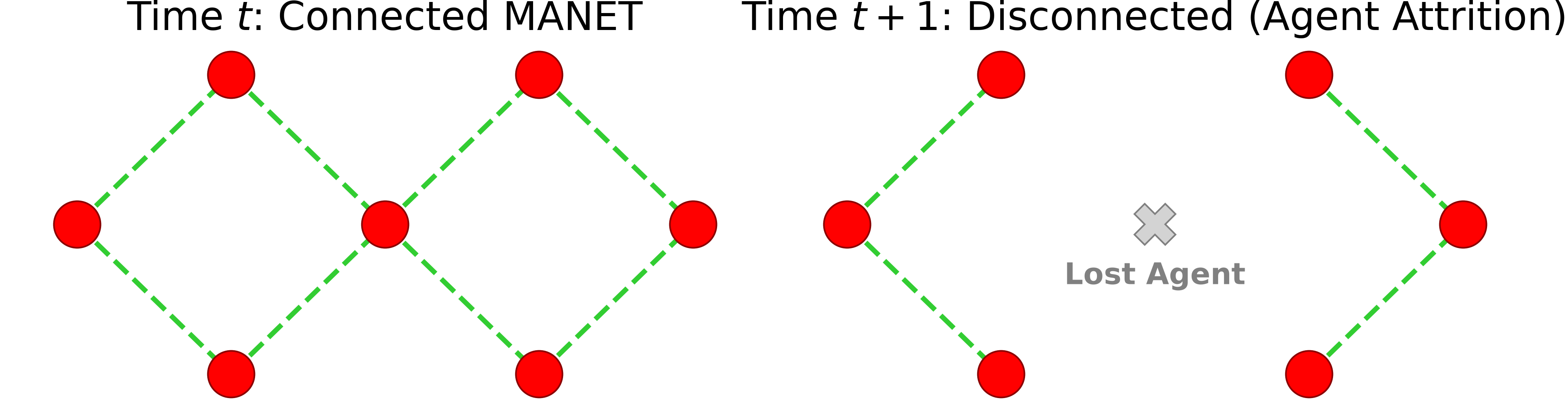}
    \caption{Graph-based representation of a disconnected MANET due to agent attrition.}
    \label{fig:agent_loss}
\end{figure}

Not only can this result in a loss of the collective knowledge of the environment due to reduced sensor nodes, but the removal of even a single agent can isolate components of the MANET, which want to work collaboratively. As illustrated in Figure \ref{fig:agent_loss}, at time $t$ the MANET has a fully connected visibility graph. At time $t+1$, the loss of a central routing node fractures the network into two disjoint components.

\begin{definition}[World Change]\label{def:world_change}
A \textit{World Change} (or mutation) is defined as an instantaneous alteration to the topological structure of the environment.
Let $W_t$ represent the navigable space at a given discrete time step $t$. A structural mutation that transitions the environment from $W_t$ to $W_{t+1}$ can be described by the addition of new free space (added floor) or the removal of existing free space (added walls).
$$W_{t+1} = (W_t \cup \Delta W_{add}) \setminus \Delta W_{sub}$$

\end{definition}
These mutations can create potentially unseen regions or introduce new line-of-sight barriers, which can sever existing edges in the visibility graph and/or, in the worst case, force a localized agent loss. Within this study, to ensure complete, connected coverage is always possible, it is assumed that $\Delta W$ cannot create impassable pinch points or disconnect the world structure. Additionally, it is assumed that the deployment point $x_d$ remains safe from world subtractions (i.e., $x_d \notin \Delta W_{sub}$), ensuring that the MANET retains a stable anchor and adaptive recovery remains feasible.

\begin{figure}[h!]
    \centering
    \includegraphics[width=0.46\textwidth]{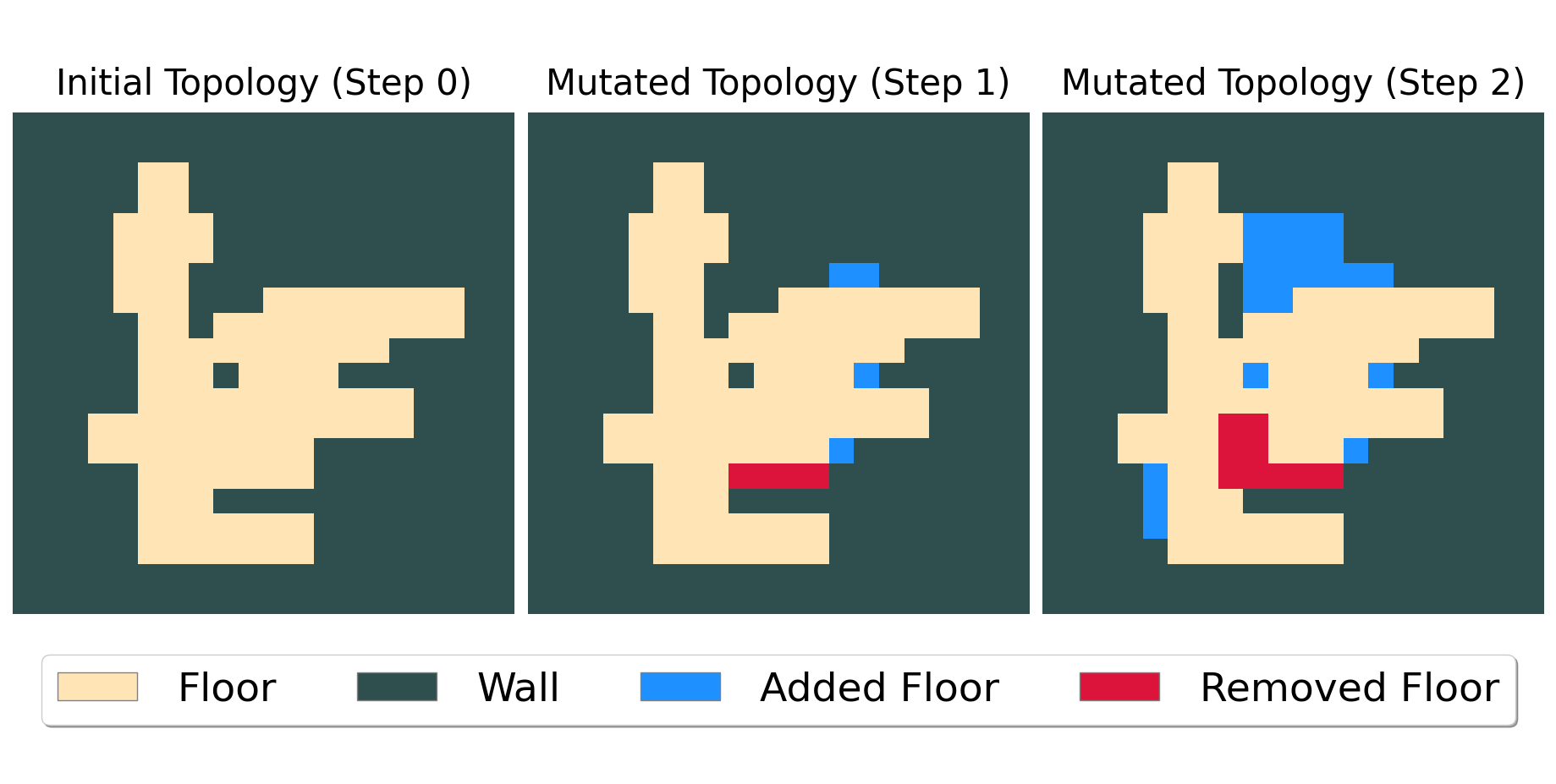}
    \caption{Example 2D dungeon map after 2 successive mutations.}
    \label{fig:mutation_ex}
\end{figure}

Both the POCGAGP and the DPOCGAGP are defined over two- and three-dimensional worlds. We
focus here on 2D orthogonal planes ($W \subset \mathbb{R}^2$), consistent with the
algorithms developed for the POCGAGP to date and with the broader AGP and CGAGP literature.
This setting captures many practical deployment scenarios and preserves the core geometric
and communication challenges while keeping computations tractable
\cite{meriaux2025manet}. Nothing in the DPOCGAGP formulation itself restricts the
disturbance models to two dimensions.

\section{Adaptive MANET-Building Algorithms}
We present advancements to CADENCE and DADENCE to adapt to the new requirements of the DPOCGAGP.
Note that in the DPOCGAGP, achieving full coverage does not end the deployment:
the MANET must remain operational for the full duration of the mission (e.g.,
the period during which users rely on the network in the area), denoted
$T_{run}$, and a disturbance may occur at any point during this time.
Both adaptive algorithms therefore run for the duration
of the mission rather than terminating once coverage is complete. When
$\mathbf{B} = \emptyset$, the network is in a steady state: agents hold their
positions and continue to monitor connectivity and coverage. Any subsequent
Agent Loss or World Change event can re-expose part of the border, and the
deployment loop resumes.
\subsection{Adaptive CADENCE}
\label{subsec:adaptive_cadence}
To modify the CADENCE algorithm to allow it to handle unpredictable disturbances, it is assumed any ``stranded" agents (i.e., those occupying a disjoint component of $G(A)$ not containing the deployment point $x_d$) are treated by the centralized algorithm as effectively destroyed until connectivity is re-established and the agent can be reincorporated. Let $A_{conn} \subseteq A$ denote the set of agents connected to $x_d$. Since
CADENCE observes the world only through $F_W(A_{conn} \cup \{x_d\})$, stranded
agents, which lie outside $A_{conn}$, are invisible to the planner.
 Additionally, it is assumed agents only contribute to the coverage ``known" by the centralized algorithm if the agent is at its terminal goal position. This avoids instability and movement into areas without guaranteed connectivity.

Three distinct recovery methods are suggested and evaluated to handle unpredictable disturbances in the case of the CADENCE algorithm. While there exist a multitude of ways to coordinate connected recovery with CADENCE coverage guarantees, these heuristics highlight three general approaches for guaranteed recovery.

\begin{enumerate}
    \item \textbf{Method 1: Redeploy to vacated valid corners.} The system draws its targets from the seen valid corners, $S'$ that are neither currently occupied nor deallocated,
    i.e.\ $S' \setminus (A_{conn} \cup A_{dealloc} \cup \{x_d\})$.

\item \textbf{Method 2: Reactively redeploy to all valid goals.} When a disturbance changes
    the edge of coverage $\mathbf{B}$, the system recomputes all valid $270^\circ$ corners in
    the known map that do not have agents deployed and re-queues them, even if previously visited. This enforces CADENCE's
    geometric coverage guarantees on the post-disturbance map, but results in high agent
    consumption.

\item \textbf{Method 3: Proactive deployment to frontier goals.} The system deploys only to
    (and never deallocates from) valid goals whose field of view contains the edge of coverage
    $\mathbf{B}$, focusing resources on expanding exploration and repairing critical gaps rather
    than redundant internal coverage. Once connectivity is regained with stranded agents, it
    will not deploy to previously explored valid corners that temporarily fell out of coverage.
\end{enumerate}

Algorithm~\ref{alg:augmented_cadence} presents the adapted CADENCE algorithm in full. The three methods differ in the candidate set they redeploy from, and this determines which disturbances they provably recover from.

Methods 2 and 3 guarantee full coverage under both \textit{Agent Loss} and \textit{World Change} because each derives its targets from the current known map rather than the pre-disturbance configuration. Method 2 recomputes the full set of valid $270^\circ$ corners on the updated map and re-queues all of them. This reduces recovery to a fresh static POCGAGP instance on the post-disturbance world, for which deploying to the complete corner set already guarantees full connected coverage~\cite{meriaux2025manet}, whether the disturbance removed agents, added floor, or added walls. Method 3 instead ties termination to the coverage border, deploying only to corners whose field of view intersects $\mathbf{B}$ and halting only when $\mathbf{B} = \emptyset$. Since any uncovered region induces a non-empty border with an incident valid corner, the loop cannot stop while coverage is incomplete. Either disturbance simply re-exposes frontier, which the loop then resolves.

\begin{figure}[h!]
    \centering
    \includegraphics[width=0.5\textwidth]{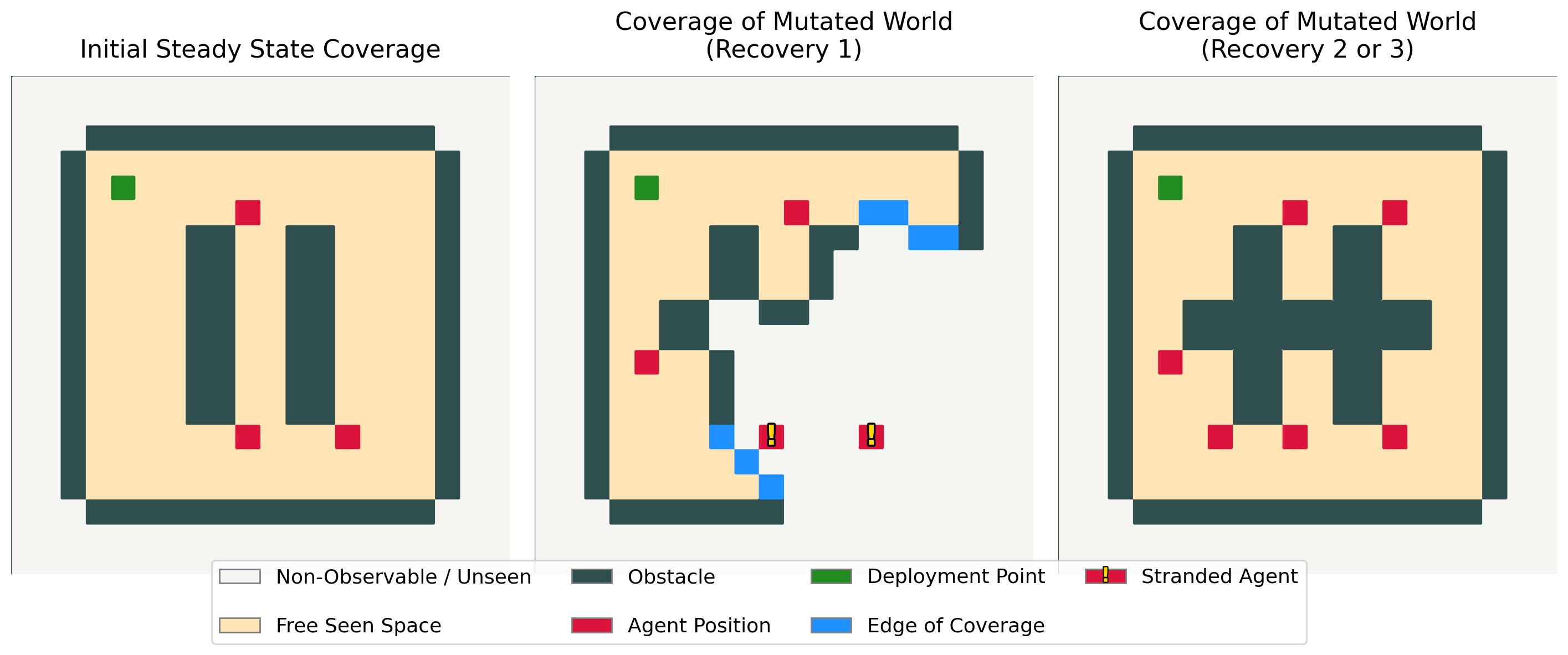}
    \caption{Steady State Coverage before (left) and after (middle, right) world mutation event.}
    \label{fig:three_images}
\end{figure}

Method 1 draws its targets from previously seen, non-deallocated corners, refilling the positions of lost agents. Under \textit{Agent Loss} this suffices, since at least one lost agent remains in view of the connected network, so its position is reachable and refilling it restores the prior covering. Under \textit{World Change} it fails for two reasons. A vacated position may no longer be a valid corner, or may no longer exist. More fundamentally, even a fully restored prior configuration need not cover a mutated world, since added floor could lie outside the view of every prior agent and Method 1 cannot target it. It therefore guarantees recovery only under \textit{Agent Loss}. Figure~\ref{fig:three_images} illustrates such a case.

\begin{algorithm}
\caption{Adaptive CADENCE}
\label{alg:augmented_cadence}
\begin{algorithmic}[1]
\State \textbf{Input:} Field of view $F_W(\cdot)$, valid corner sensor $\alpha(\cdot)$, deployment point $x_d$, recovery heuristic $\mathcal{M} \in \{1, 2, 3\}$, and mission duration $T_{run}$
\State \textbf{Initialize:}
\State \hspace{\algorithmicindent} Agent sets: deployed $A \leftarrow \emptyset$, connected $A_{conn} \leftarrow \emptyset$, deallocated $A_{dealloc} \leftarrow \emptyset$
\State \hspace{\algorithmicindent} Seen valid corners $S' \leftarrow \alpha(\{x_d\})$
\State \hspace{\algorithmicindent} Border of connected vision $\mathbf{B} \leftarrow B(F_W(\{x_d\}))$
\State \hspace{\algorithmicindent} Mission time $T \leftarrow 0$
\While{$T < T_{run}$}
    \State $A_{conn} \leftarrow$ Agents in $A$ connected to $x_d$
    \State Update $\mathbf{B} \leftarrow B(F_W(A_{conn} \cup \{x_d\}))$
    \While{$\mathbf{B} \neq \emptyset$ \textbf{and} $T < T_{run}$}
        \State $A_{conn} \leftarrow$ Agents in $A$ connected to $x_d$
        \State Update seen valid corners $S' \leftarrow \alpha(A_{conn} \cup \{x_d\})$
        \State Update $\mathbf{B} \leftarrow B(F_W(A_{conn} \cup \{x_d\}))$
        \If{world change \textbf{or} new agent loss detected}
            \If{heuristic $\mathcal{M} = 2$}
                \State Reset deallocation history $A_{dealloc} \leftarrow \emptyset$
            \EndIf
        \EndIf
        \If{heuristic $\mathcal{M} \in \{1, 2\}$}
            \State Pick $x \in S' \setminus (A_{conn} \cup A_{dealloc} \cup \{x_d\})$
        \ElsIf{heuristic $\mathcal{M} = 3$}
            \State Pick $x \in (\alpha(\mathbf{B}) \cap S') \setminus (A_{conn} \cup \{x_d\})$
        \EndIf
        \State Deploy new agent to $x$: $A \leftarrow A \cup \{x\}$
        \State Sequentially transfer agents from $A_{conn}$ to $A_{dealloc}$ if their removal preserves both connectivity and coverage. Then remove the same agents from $A$.
    \EndWhile
\EndWhile
\end{algorithmic}
\end{algorithm}

\subsection{Adaptive DADENCE}

DADENCE is itself not optimized for agent recovery; to address this, we propose to improve the algorithm in the following manner in Algorithm~\ref{alg:dadence_decentralized}.

\begin{figure}[h!]
    \centering
    \includegraphics[width=0.45\textwidth]{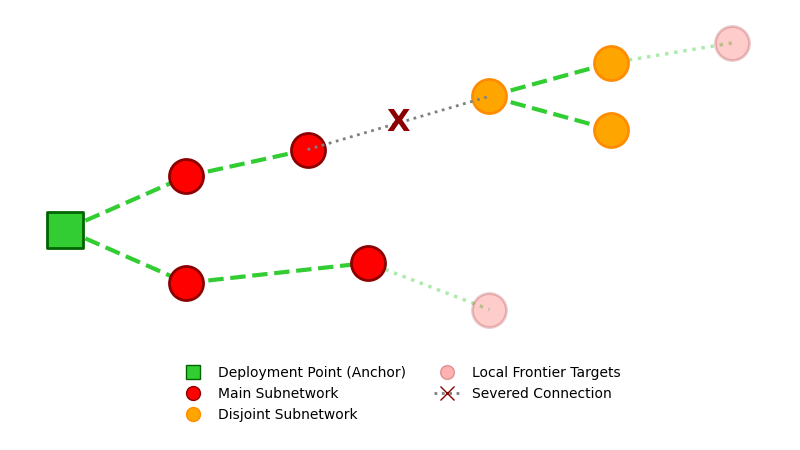}
    \caption{Graph-based representation of an example of network fragmentation and subnetwork independence.}
    \label{fig:network_fragment}
\end{figure}

When an \textit{Agent Loss} or \textit{World Change} event severs a communication link, the visibility graph fractures into two or more disjoint components, referred to as subnetworks. Because DADENCE agents operate entirely on local communication and shared observations within their connected component, each fractured subnetwork naturally transitions to operating as an independent swarm. As illustrated in Figure \ref{fig:network_fragment}, a severance creates a main subnetwork that retains its connection to the deployment point (anchor), alongside one or more disjoint subnetworks.

Following a fracture, the agents within each isolated subnetwork pool their localized field of view and independently calculate a new local frontier target, $x^*$. The main subnetwork connected to $x_d$ continues its normal operation, utilizing its access to the anchor to spawn replacement agents, fill coverage gaps, and expand towards its own calculated frontier. The isolated subnetworks, which no longer have a routing path back to the deployment point $x_d$, continue to explore and expand their coverage autonomously, the only difference being that they are unable to spawn new agents to replace those lost and allow for continuous exploration. These disjoint swarms are effectively ``stuck'' with their current agent count,  using only their surviving nodes until they reach a static equilibrium.

When the independent exploration of these disjoint swarms naturally leads to a renewed line-of-sight connection, the subnetworks merge. Upon reconnection, the distinct local goals of each subgroup are immediately discarded. The newly merged network synchronizes its combined discrete field of view and collaboratively calculates a unified frontier target to resume exploration.

\subsection{Adaptive ISDA (Benchmark)}
A centralized Incremental Self-Deployment Algorithm (ISDA) \cite{howard2002ISDA} is used to benchmark the performance of the adaptive recovery of CADENCE and DADENCE. ISDA incrementally deploys a
MANET by placing new agents at randomly selected points along the edge $\mathbf{B}$ of the
current coverage area. Similar to DADENCE, the ISDA framework naturally extends to the dynamic setting; the algorithm requires no memory of previous agent positions, so any border exposed by a disturbance is organically included in each subsequent frontier goal computation. Similarly, ISDA can be easily extended to adopt the  deallocation policy described in Section~\ref{subsec:deallocation}.

\begin{algorithm}
\caption{Adaptive DADENCE (Runs in parallel on each Agent $a$)}
\label{alg:dadence_decentralized}
\begin{algorithmic}[1]
\State \textbf{Input:} Agent $a$, Field of View $F_W(\cdot)$, deployment point $x_d$, and mission duration $T_{run}$
\State \textbf{Initialize:}
\State \hspace{\algorithmicindent} Previous connected subnetwork $A_{a}^- \leftarrow \emptyset$
\State \hspace{\algorithmicindent} $A_a \leftarrow$ All agents connected to agent $a$, including $x_d$ (a static ``pseudo'' agent) if it belongs to the same subnetwork
\State \hspace{\algorithmicindent} Local border $\mathbf{B}_a \leftarrow B(F_W(A_a))$
\State \hspace{\algorithmicindent} Mission time $T \leftarrow 0$
\While{$T < T_{run}$}
    \State Update $A_a$ and $\mathbf{B}_a \leftarrow B(F_W(A_a))$
    \While{$\mathbf{B}_a \neq \emptyset$ \textbf{and} $T < T_{run}$}
        \State Update $A_a$
        \State Update subnetwork vision $\mathbf{B}_a \leftarrow B(F_W(A_a))$
        \If{$x_d \in A_a$ \textbf{and} $A_a = A_{a}^-$}
            \State Send deployment request to $x_d$
        \Else
            \State $A_{a}^- \leftarrow A_a$
            \State Collaboratively choose frontier target $x^* \in \mathbf{B}_a$
            \State Find next step $a^+$ along shortest path to $x^*$
            \State Broadcast $a^+$ to $A_a$ and receive their steps
            \If{proposed moves of $A_a$ maintain connectivity and coverage \textbf{and} avoid collisions}
                \State Take action and execute move: $a \leftarrow a^+$
            \Else
                \State Remain at current position
            \EndIf
        \EndIf
    \EndWhile
\EndWhile
\end{algorithmic}
\end{algorithm}

\section{Implementation and Testing}
\subsection{World Generation}
To enable fair comparisons between unique orthogonal and planar worlds, an environmental pipeline was created to normally distribute the size ($S$) and number  ($N_f$) of unique world features (termed rooms, hallways, and obstacles), and uniformly distribute the spatial position $P$ of each feature.
\begin{equation*}
    N_{f}, S \sim \mathcal{N}(\mu, \sigma^2), \quad P \sim \mathcal{U}(P_{min}, P_{max})
\end{equation*}
where $\mu$ and $\sigma$ are parameterized as linear functions of the map dimensions (i.e., width and height).

In the case of \textit{World Change}, a sequence of world states $\{W_0, W_1, \dots, W_m\}$ were then generated by applying the same spatial logic to a mutation step, $\Delta W$. An example of $\mathbb{R}^2$ worlds generated from this pipeline, with the initial topology and subsequent mutations, is shown in Figure \ref{fig:mutation_ex}.

\subsection{Single World Agent Loss Example}
To better understand the resource usage during adaptive recovery, simulations on a single, randomly sampled $30 \times 30$ grid world were conducted, with a single loss event occurring once all algorithms had achieved full world coverage for the first time. The results highlight the trade-offs between the recovery methods in terms of the robustness metrics described earlier. Note that these results  represent performance on one specific map --- and that the loss event does not affect each simulation equally depending on the network configuration at the time of loss.

\begin{figure}[h!]
    \centering
    \begin{subfigure}[b]{\linewidth}
        \centering
        \begin{minipage}[c]{1.5em}
            \caption{}
            \label{fig:q9_step}
        \end{minipage}%
        \begin{minipage}[c]{0.784\linewidth}
            \includegraphics[width=\linewidth]{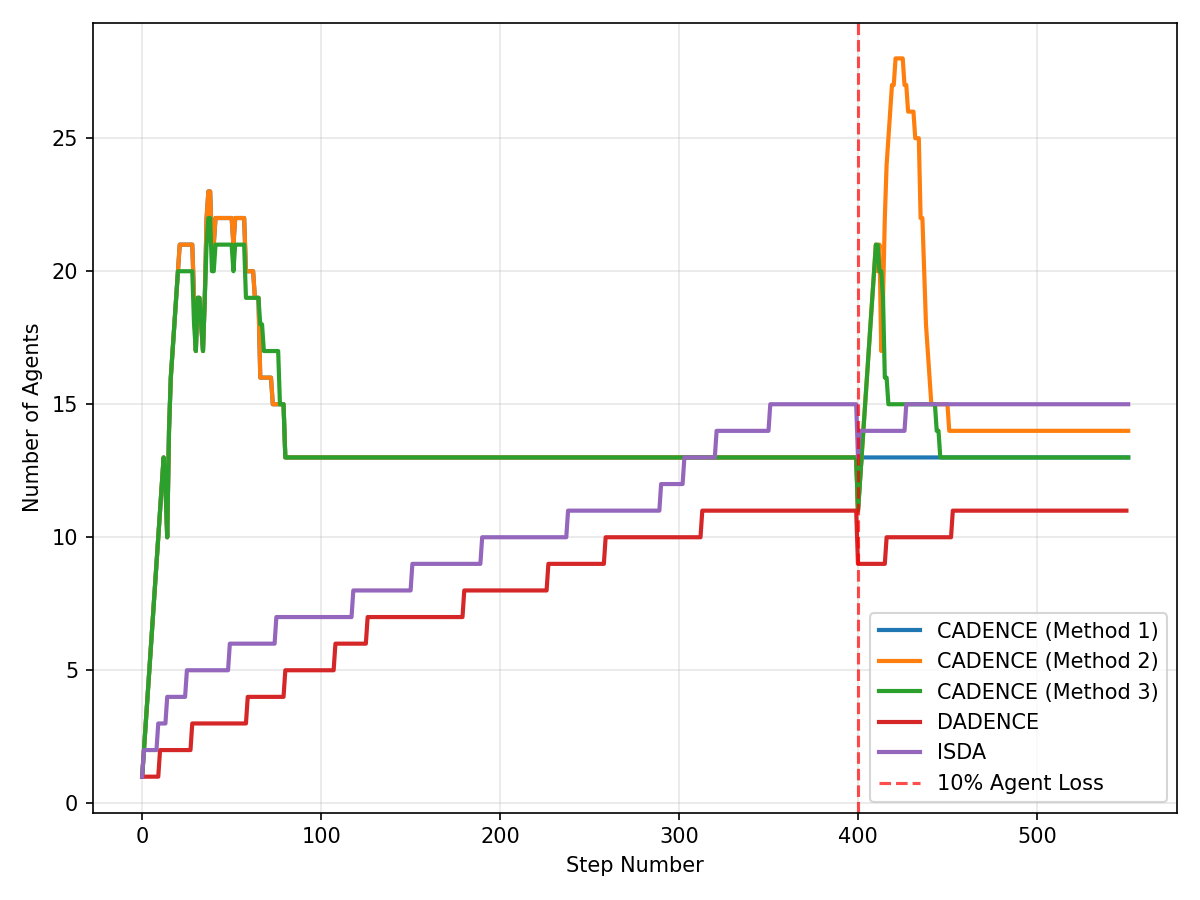}
        \end{minipage}
    \end{subfigure}

    \vspace{0.5em}

    \begin{subfigure}[b]{\linewidth}
        \centering
        \begin{minipage}[c]{1.5em}
            \caption{}
            \label{fig:q9_good_traj}
        \end{minipage}%
        \begin{minipage}[c]{0.784\linewidth}
            \includegraphics[width=\linewidth]{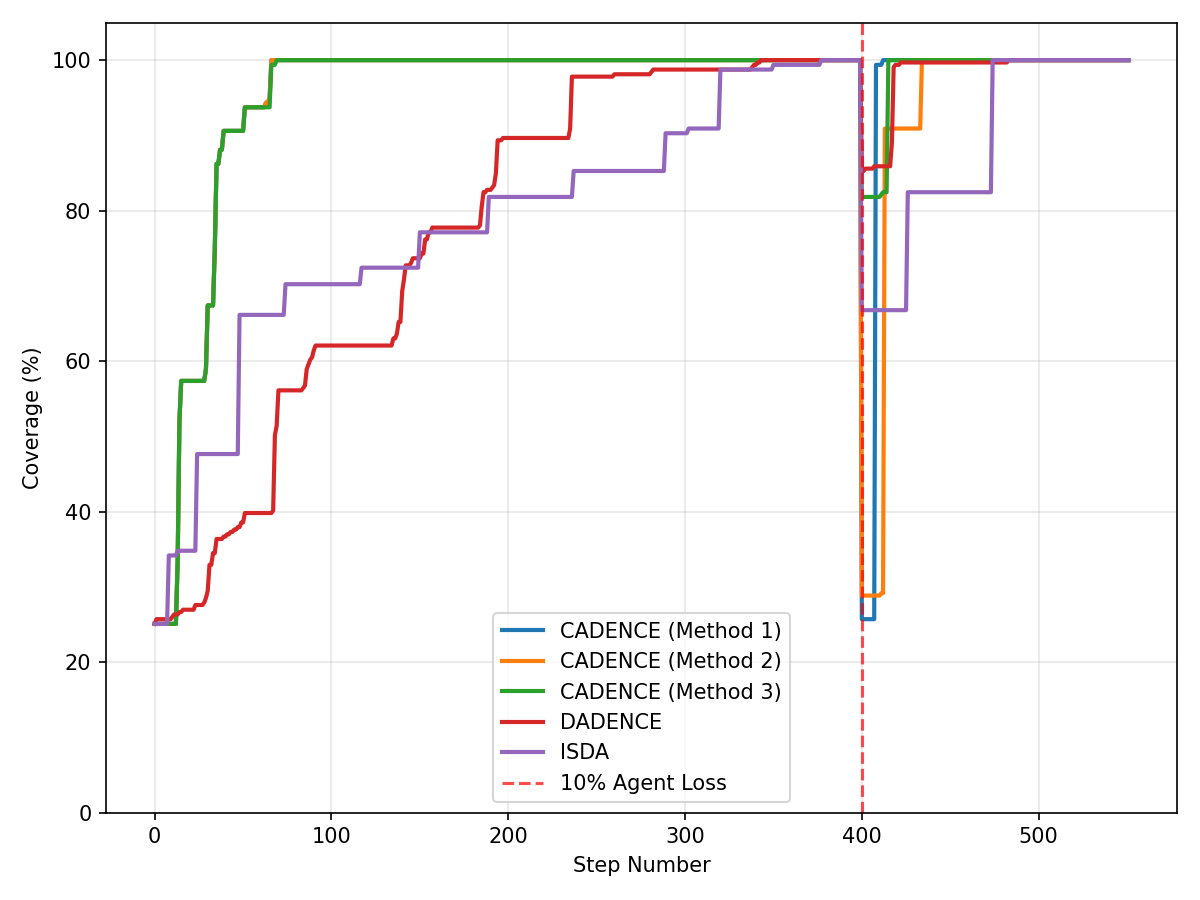}
        \end{minipage}
    \end{subfigure}

    \caption{Comparison of agent use and map coverage for an example
    $30 \times 30$ world with a 10\% agent loss at step 400: (a) active agents
    over time; (b) map coverage (\%) over time.}
    \label{fig:q9_combined}
\end{figure}

Before evaluating the disturbance, the initial deployment phase (Steps 0 to 400) clearly demonstrates the operational differences between the original algorithms. CADENCE rapidly achieves 100\% coverage by aggressively deploying agents, while DADENCE prioritizes agent efficiency; it maintains a steady, low active agent count but requires more steps to fully explore the environment.
As seen in Figures \ref{fig:q9_step} and \ref{fig:q9_good_traj}, a 10\% agent loss at step 400 instantly fractures the network and drops coverage. DADENCE recovers at a slightly slower rate than CADENCE but still achieves a quick recovery with conservative agent use.  DADENCE outperforms ISDA in both metrics.

\subsection{Numerical Testing Methodology}

\subsubsection{Preprocessing for Large Scale Tests}
To evaluate the adaptive algorithms against a wide distribution of topological challenges, 200 worlds of varying sizes and complexities were generated~\cite{meriaux2026dungeon} (100 for agent loss and 100 for world change). For each generated world, 5 distinct deployment points were assigned using 5 fixed random seeds, establishing 500 unique initial states per case (1000 total). To maintain a controlled experimental baseline, all random parameters, including the coordinates of the 5 deployment points, were generated exactly once per world. These fixed parameters were then applied universally across all algorithms tested on that specific map.

Each simulation undergoes 5 sequential disturbance events during execution. For the sake of consistency, the time of each disturbance is linked to the progression of the exploration rather than an arbitrary time step. An event is triggered dynamically whenever the algorithm successfully covers a randomly generated percentage of the remaining uncovered free space. This target coverage threshold is recalculated relative to the state of the map at the previous disturbance event. In \textit{Agent Loss}, each of these 5 events corresponds to a random 10\% loss of the currently active agents in the network.

As with the deployment points, the exact percentage thresholds triggering the 5 change events are generated once for each world seed from a uniform distribution.
It is important to acknowledge that a perfectly ``fair" one-to-one comparison of a disturbance event is functionally impossible in DPOCGAGP. Because these algorithms can deploy agents at vastly different rates and form entirely different spatial configurations, triggering a 10\% agent loss at the exact same coverage threshold will inherently impact a sparse network differently than a dense network.
By aggregating the performance data across 200 distinct worlds and 1000 unique deployment trajectories, the impact of situational variances between disturbance events is reduced, allowing generalized performance trends to emerge.

Each initial state was evaluated with every applicable algorithm. Under
\textit{Agent Loss}, five algorithms were tested (CADENCE Methods 1--3,
DADENCE, and ISDA), giving $2500$ simulations. Under
\textit{World Change}, CADENCE Method 1 was excluded (see Section~\ref{subsec:adaptive_cadence}), leaving four algorithms and $2000$ simulations, for a total of 4500
simulations.

\subsubsection{Map Complexity}
\begin{figure}[htbp]
    \centering
    \begin{subfigure}[c]{0.24\textwidth}
        \centering
        \includegraphics[width=\linewidth]{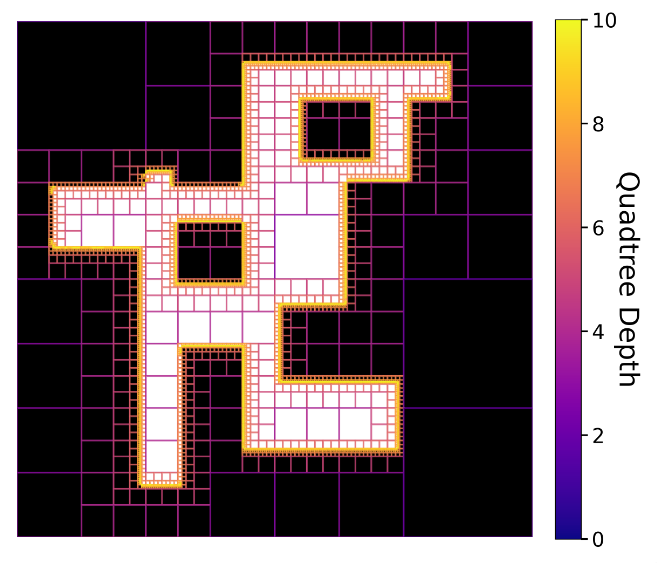}
        \caption{Quadtree decomposition of an example world map \cite{meriaux2025manet}}
        \label{fig:quadtree}
    \end{subfigure}
    \hfill
    \begin{subfigure}[c]{0.24\textwidth}
        \centering

        \begin{subfigure}{\linewidth}
            \centering
            \includegraphics[width=\linewidth]{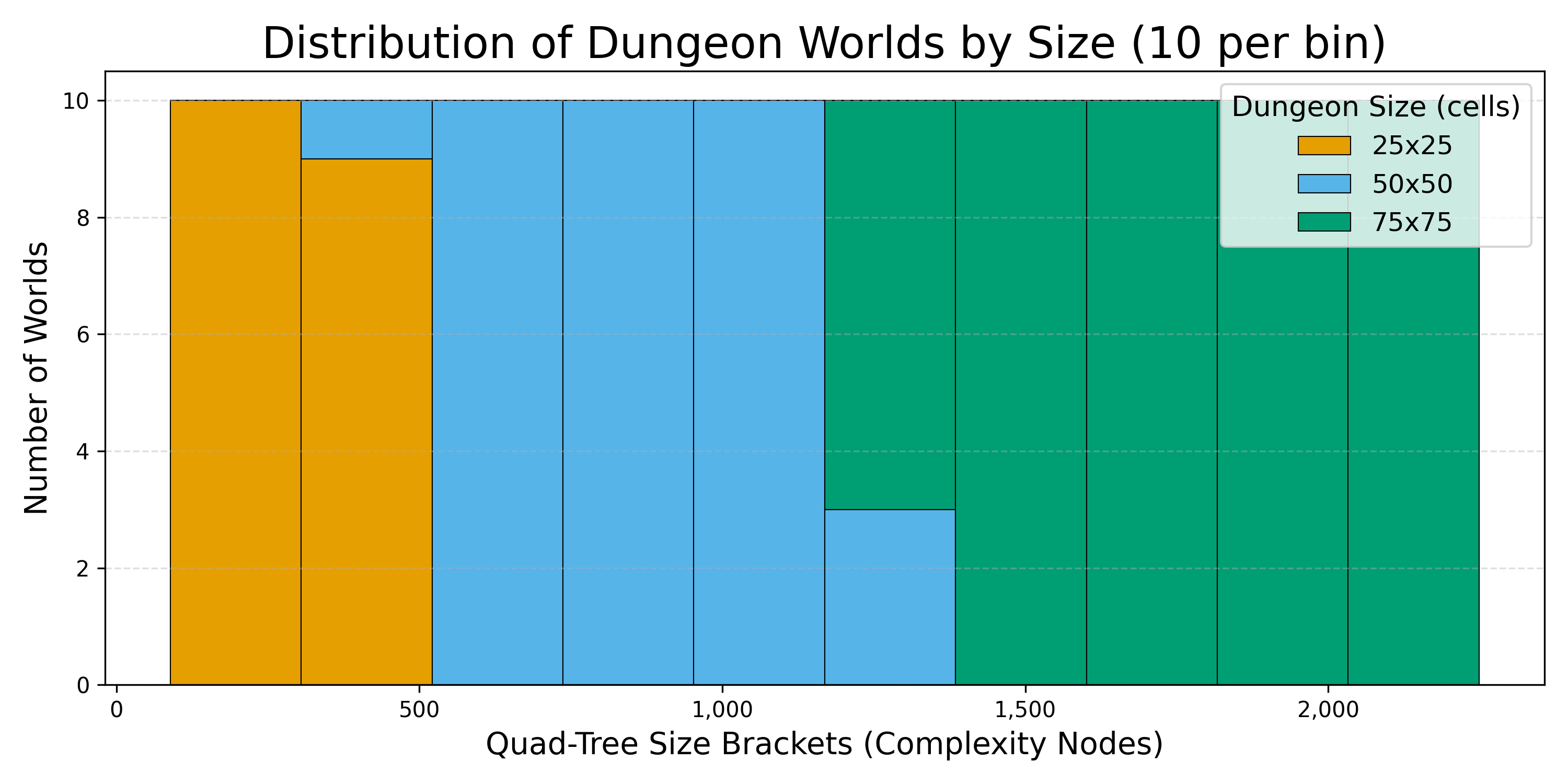}
            \caption{Agent Loss World Complexity Histogram}
            \label{fig:complexity}
        \end{subfigure}

        \vspace{0.5cm}

        \begin{subfigure}{\linewidth}
            \centering
            \includegraphics[width=\linewidth]{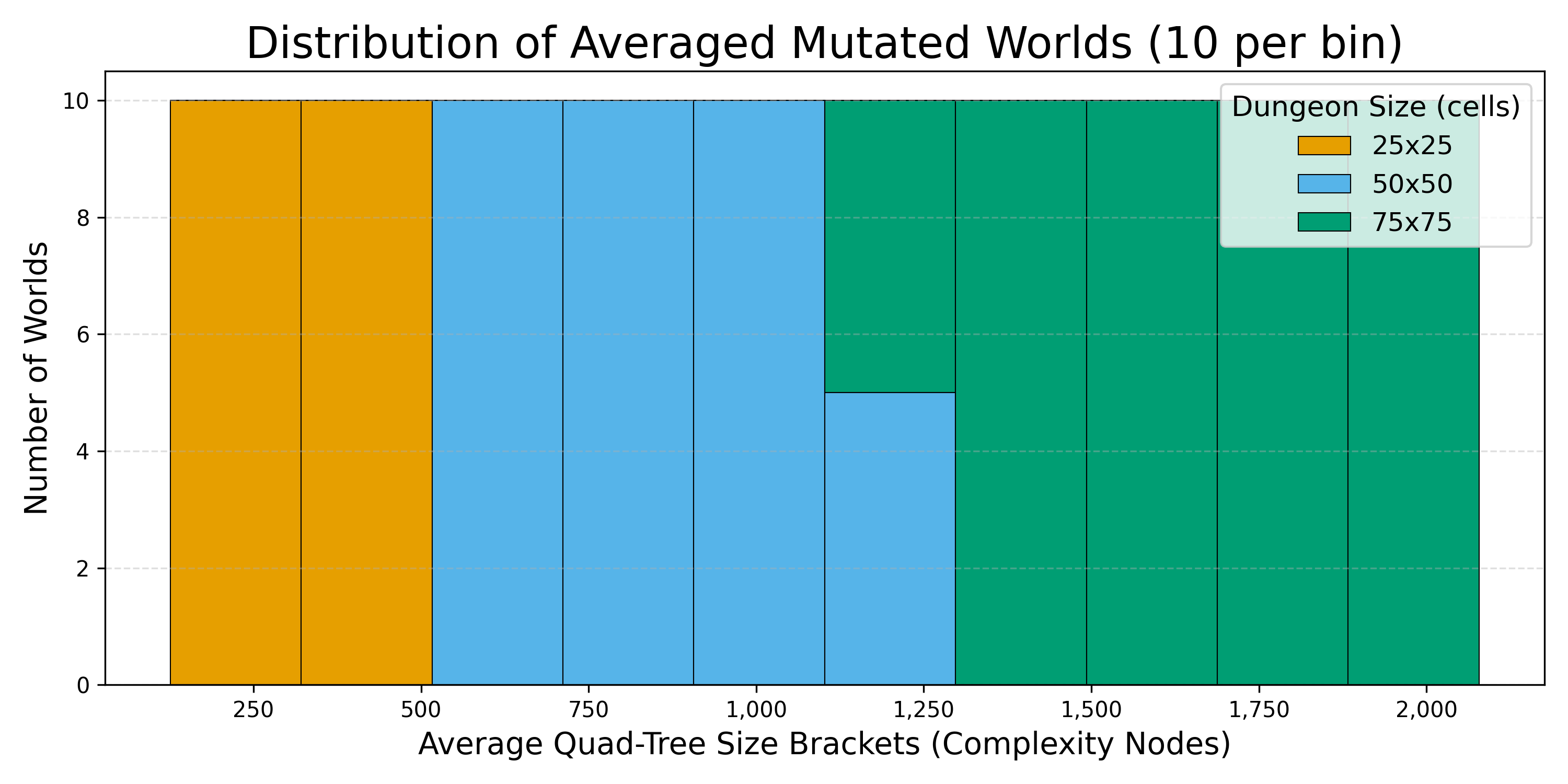}
            \caption{World Mutation Average World Complexity Histogram}
            \label{fig:mutation}
        \end{subfigure}

    \end{subfigure}

    \caption{Example quadtree decomposition and world complexity bin distributions for 100 worlds}
    \label{fig:comparison}
\end{figure}

To accurately quantify the spatial difficulty of each generated environment, map complexity is measured using a quadtree decomposition, as illustrated in Figure \ref{fig:quadtree} (see \cite{meriaux2025manet} for details).
Once a complexity score is assigned, the worlds can then be categorized, as detailed in Figures \ref{fig:complexity} and \ref{fig:mutation}, and the results for each bracket can be aggregated. To handle the different world topologies seen in the world mutation case, the average quadtree complexity is computed across the original world and its 5 mutated topologies.

\subsubsection{Resource Bounds}
For each world, $N_{max}$ is set to the number of valid corners, i.e., the bound for which CADENCE guarantees connected coverage
\cite{meriaux2025manet}; under \textit{World Change}, it is recomputed on the
current world $W_t$ after each mutation. $T_{max}$ is set to the map area
($w \times h$ cells) and applies to each recovery phase, resetting after every
disturbance event.
These bounds are not proven to hold for every DPOCGAGP instance, but all
simulations were verified to remain within them.

\subsection{Results}

The results across the 1000 randomly generated world seeds confirm that both the adaptive
CADENCE heuristics and adaptive DADENCE outperform the ISDA benchmark, though the nature of
the improvement differs between the two architectures.

\begin{figure}[h!]
    \centering

    \begin{subfigure}[b]{\linewidth}
        \begin{minipage}[c]{0.08\linewidth}
            \caption{}
            \label{fig:loss_all_agent}
        \end{minipage}%
        \begin{minipage}[c]{0.91\linewidth}
            \includegraphics[width=\linewidth]{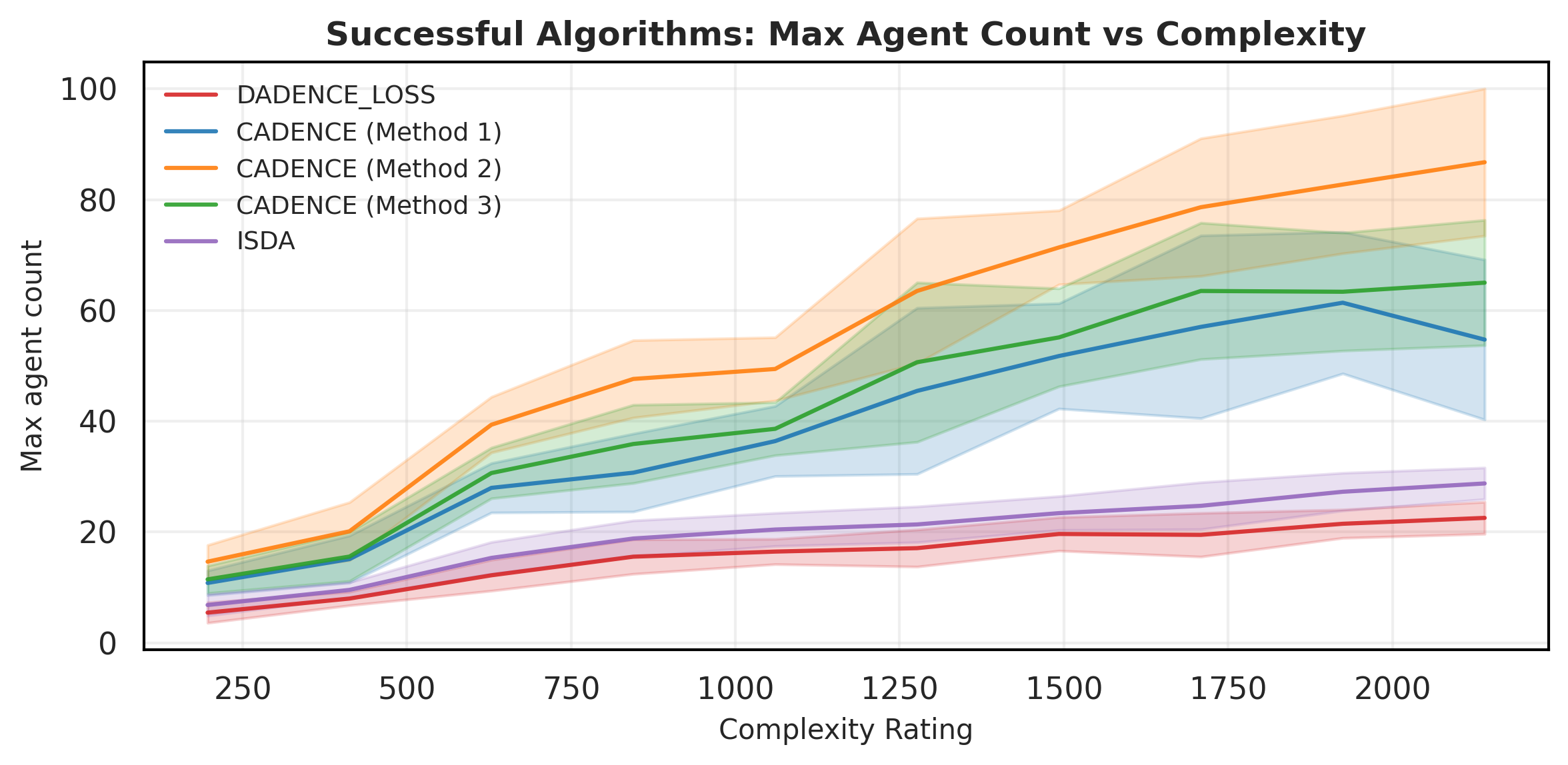}
        \end{minipage}
    \end{subfigure}

    \vspace{0.5em}

    \begin{subfigure}[b]{\linewidth}
        \begin{minipage}[c]{0.08\linewidth}
            \caption{}
            \label{fig:loss_all_final_agent}
        \end{minipage}%
        \begin{minipage}[c]{0.91\linewidth}
            \includegraphics[width=\linewidth]{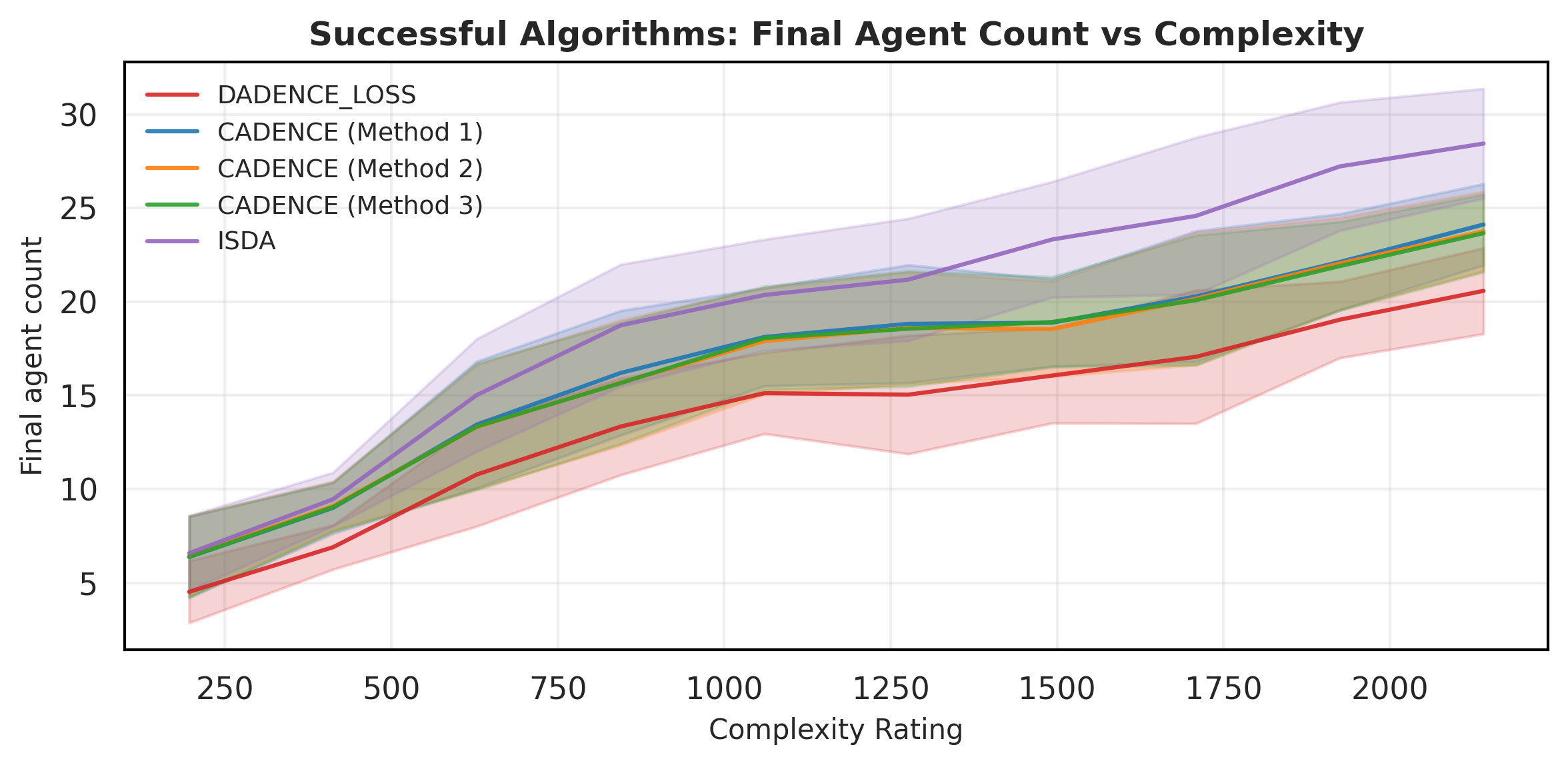}
        \end{minipage}
    \end{subfigure}

    \vspace{0.5em}

    \begin{subfigure}[b]{\linewidth}
        \begin{minipage}[c]{0.08\linewidth}
            \caption{}
            \label{fig:loss_all_steps}
        \end{minipage}%
        \begin{minipage}[c]{0.91\linewidth}
            \includegraphics[width=\linewidth]{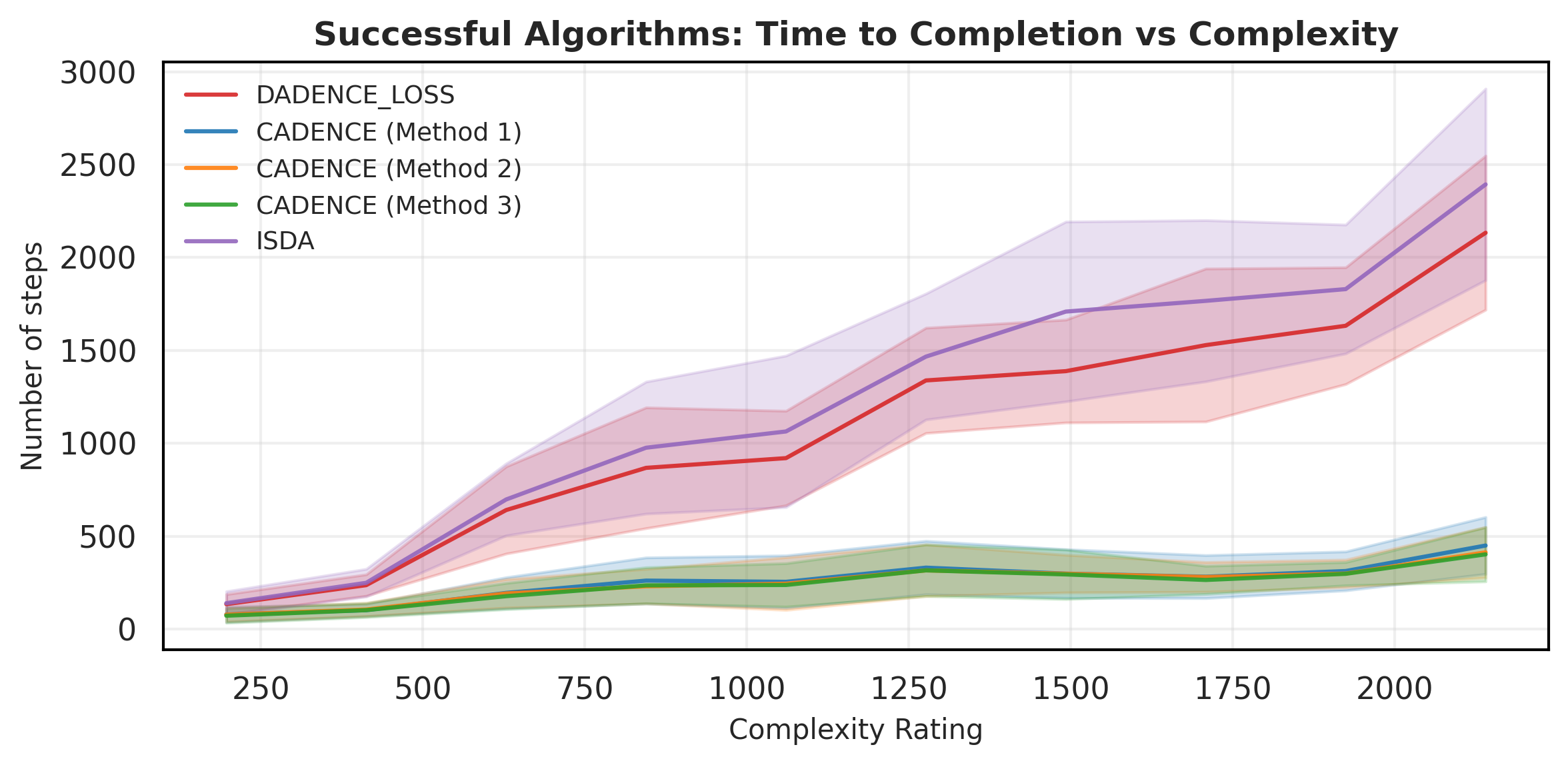}
        \end{minipage}
    \end{subfigure}

    \caption{Performance metrics for 500 world seed tests with 5 stochastic 10\% agent loss events against complexity: (a) Max Agent Use; (b) Final Agent Count; (c) Final Step Count.}
    \label{fig:combined_loss_graphs}
\end{figure}

\begin{figure}[h!]
    \centering

    \begin{subfigure}[b]{\linewidth}
        \begin{minipage}[c]{0.08\linewidth}
            \caption{}
            \label{fig:world_change_agent}
        \end{minipage}%
        \begin{minipage}[c]{0.91\linewidth}
            \includegraphics[width=\linewidth]{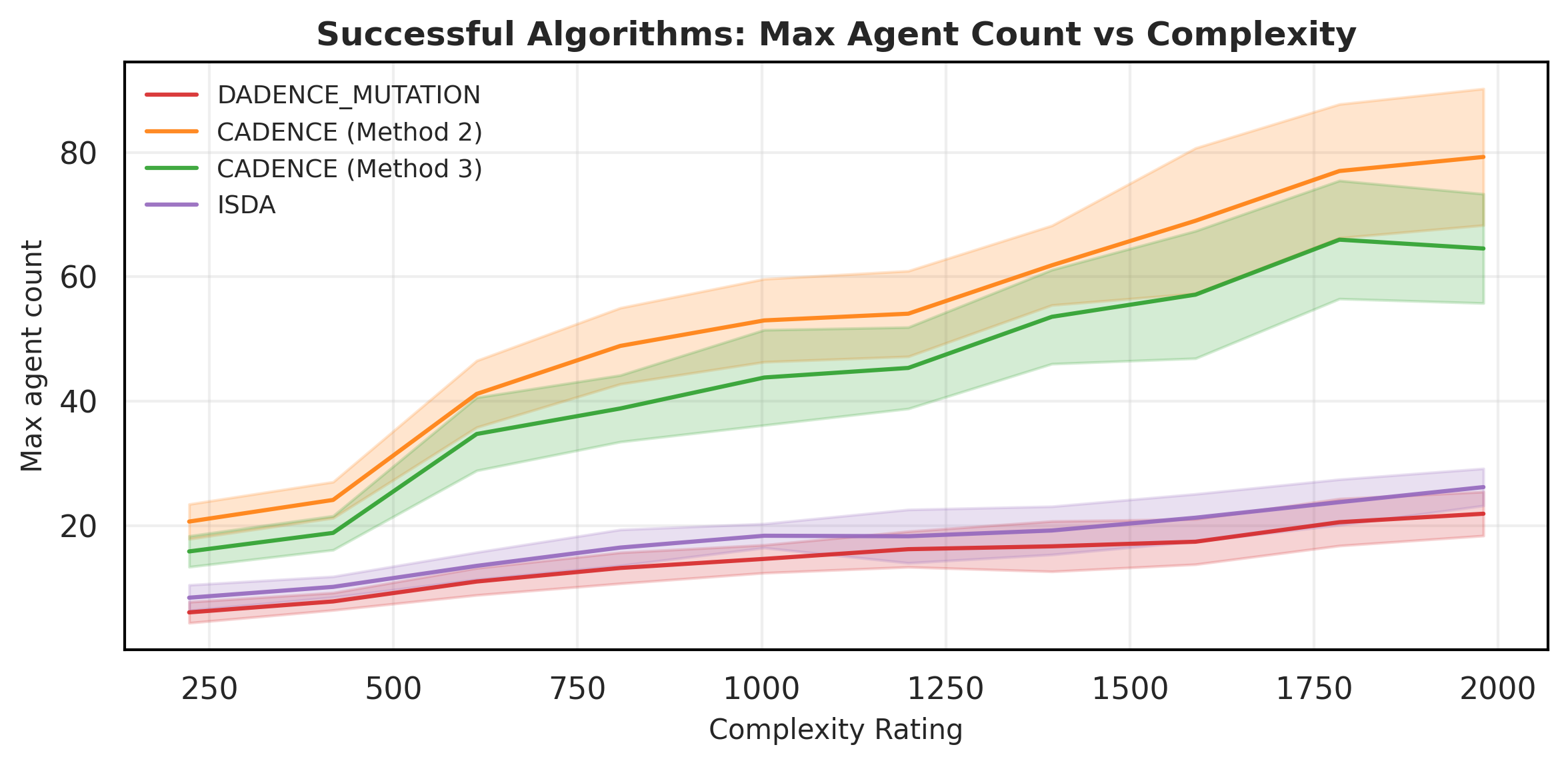}
        \end{minipage}
    \end{subfigure}

    \vspace{0.5em}

    \begin{subfigure}[b]{\linewidth}
        \begin{minipage}[c]{0.08\linewidth}
            \caption{}
            \label{fig:world_change_all_final_agent}
        \end{minipage}%
        \begin{minipage}[c]{0.91\linewidth}
            \includegraphics[width=\linewidth]{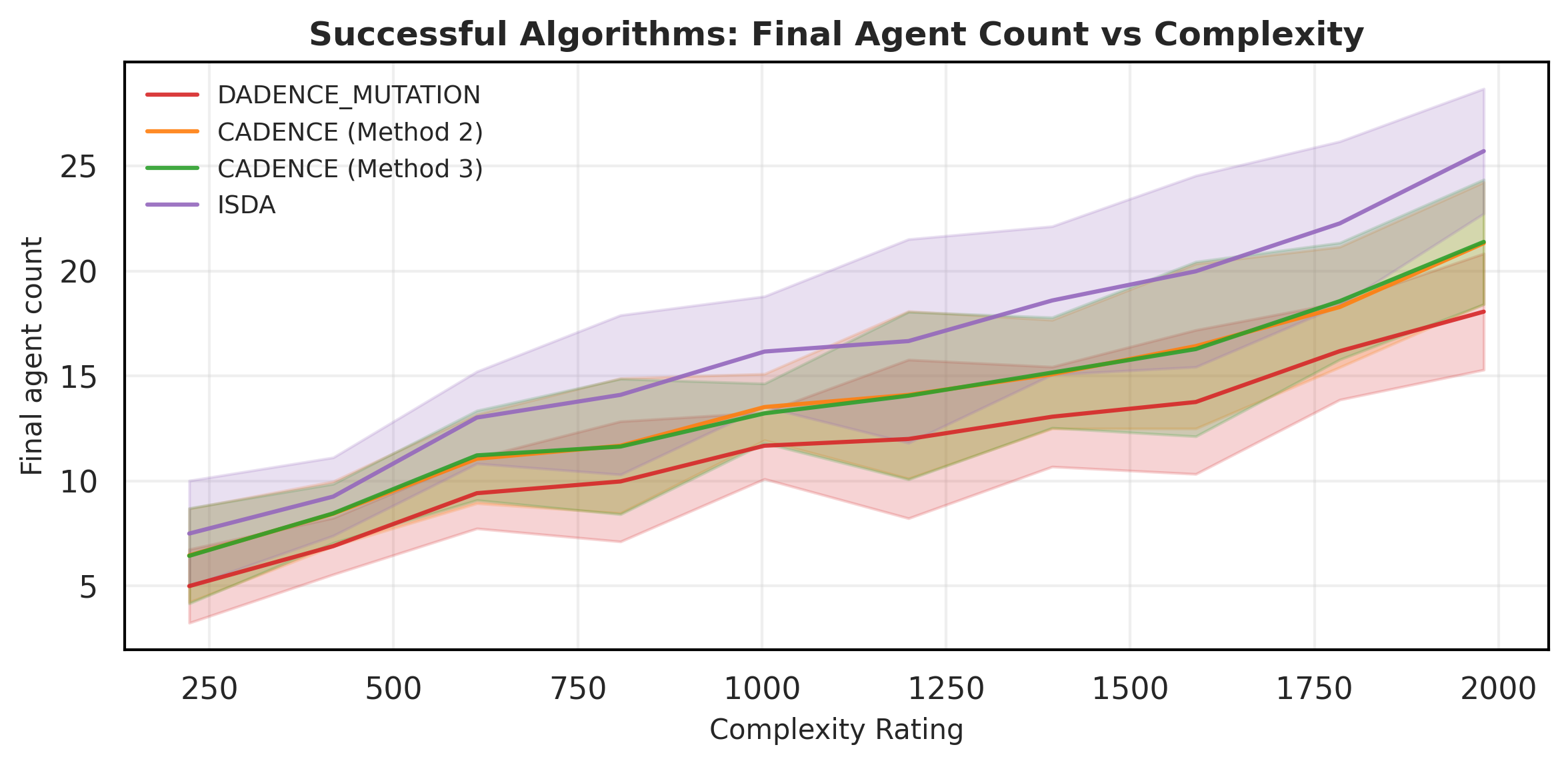}
        \end{minipage}
    \end{subfigure}

    \vspace{0.5em}

    \begin{subfigure}[b]{\linewidth}
        \begin{minipage}[c]{0.08\linewidth}
            \caption{}
            \label{fig:world_change_steps}
        \end{minipage}%
        \begin{minipage}[c]{0.91\linewidth}
            \includegraphics[width=\linewidth]{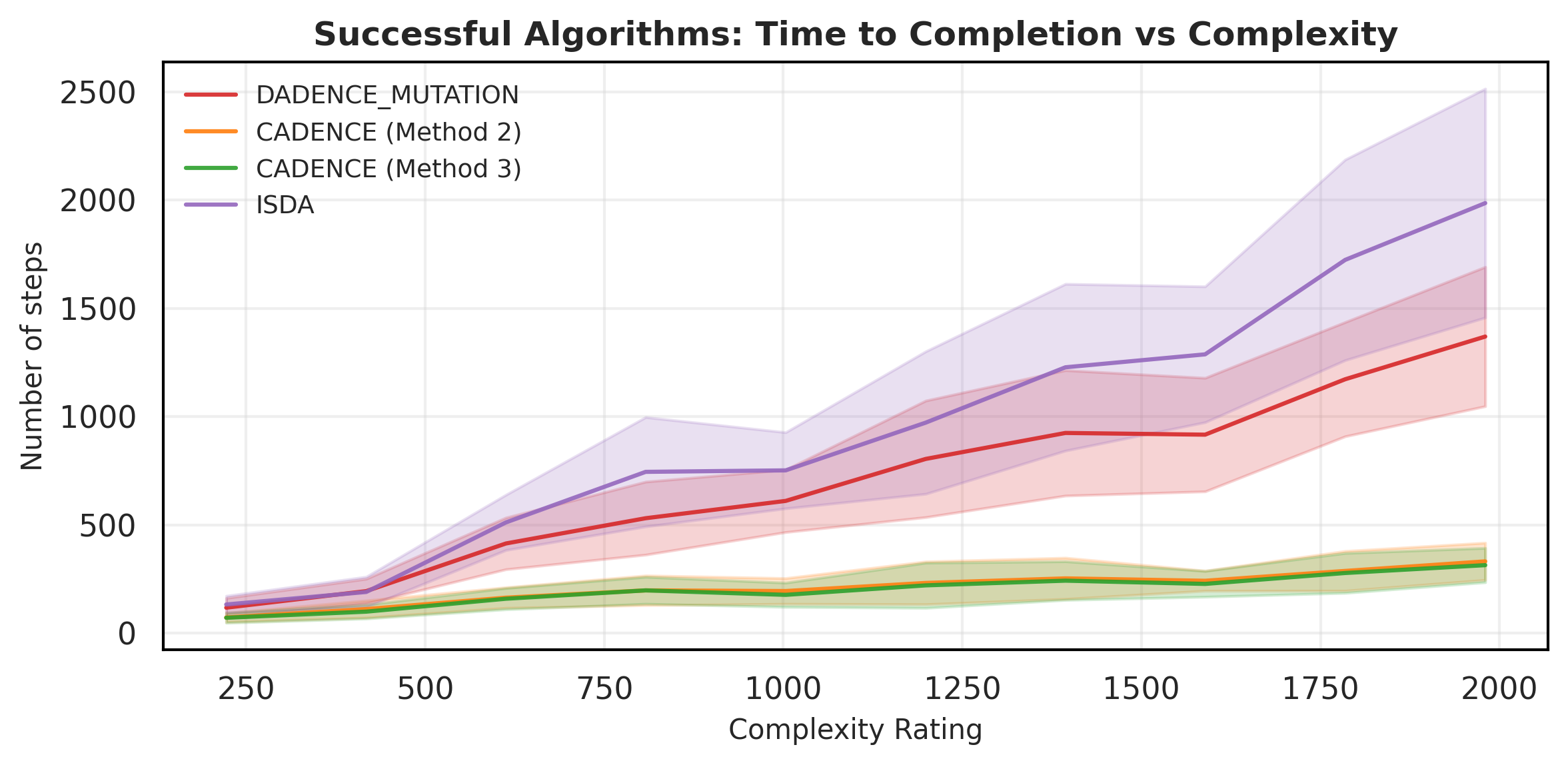}
        \end{minipage}
    \end{subfigure}

    \caption{Performance metrics for 500 world seed tests with 5 stochastic world changes against complexity: (a) Max Agent Use; (b) Final Agent Count; (c) Final Step Count.}
    \label{fig:combined_world_change_graphs}
\end{figure}

\textbf{Recovery speed.} Figures \ref{fig:loss_all_steps} and \ref{fig:world_change_steps}
show that all adaptive methods restore full coverage in fewer steps than ISDA, with the gap
widening as map complexity increases. The centralized CADENCE heuristics are the clear
leaders on this axis: by moving each agent directly to a precomputed valid-corner goal, they
require 78.88\% fewer steps than ISDA on average, and 74.23\% fewer than DADENCE. The three CADENCE
variants are nearly indistinguishable in step count (within 7\% of each other), and
their step counts grow only modestly with complexity, from roughly 75 steps in the
lowest bracket to 300--450 steps in the highest. DADENCE sits between the two: it requires
18.06\% fewer steps than ISDA on average. While DADENCE and ISDA perform similarly at the
lowest complexity brackets, the gap widens to 10.86\% in the highest complexity scenarios
under \textit{Agent Loss} and to a pronounced 31.03\% under \textit{World Change}. This
stronger scaling in mutated worlds is likely due to DADENCE's decentralized nature, which
avoids ``overreacting'' to localized topological changes: if a structural mutation only
slightly affects the edge of coverage, the nearby network simply shifts to accommodate.

\textbf{Agent efficiency.} The speed advantage of CADENCE is not free. As shown in Figures
\ref{fig:loss_all_agent} and \ref{fig:world_change_agent}, CADENCE achieves its rapid
recovery by deploying substantially larger agent pools, with peak agent counts
151.7\% higher than ISDA and 3.1$\times$ those of DADENCE on average. DADENCE is the
most agent-efficient of all approaches, requiring 18.48\% fewer maximum agents than ISDA and
avoiding the agent surges that CADENCE relies on. After deallocation, however, the final
steady-state agent counts of both CADENCE and DADENCE are lower than those of ISDA (by
13.85\% and 28.94\%, respectively; Figures \ref{fig:loss_all_final_agent} and
\ref{fig:world_change_all_final_agent}). For CADENCE, deallocation removes 67.1\% of the
peak deployment, indicating that the surge is transient: valid-corner placement yields
compact final configurations, but reaching them quickly requires temporarily
over-deploying. Across the three CADENCE variants, final agent counts after deallocation are
nearly indistinguishable (Figures~\ref{fig:loss_all_final_agent} and~\ref{fig:world_change_all_final_agent}), whereas peak agent counts diverge
as complexity increases, with Method~2 consistently the highest due to
re-queuing all valid corners after each disturbance (Figures~\ref{fig:loss_all_agent} and~~\ref{fig:world_change_agent}).

\textbf{Summary.} A clear division of strengths emerges, consistent with
\cite{meriaux2025manet}: CADENCE is best in step efficiency, DADENCE in agent efficiency,
and ISDA is beaten on both metrics by at least one adaptive method under every disturbance
type, leading in none. The system response to \textit{World Change} shows the same
performance divides as \textit{Agent Loss}, with the gaps between methods amplified.

\section{Conclusion}

This study formally introduced the DPOCGAGP to address the realities of dynamic instability in MANET deployments. By exposing both centralized and decentralized deployment algorithms to simulated agent attrition and environmental mutations across a large scale of procedurally generated environments, a fundamental operational trade-off was identified over the 4500 test cases. Centralized approaches, specifically the adaptive CADENCE frontier heuristic, proved highly effective at rapidly repairing fractured communication graphs and reclaiming lost coverage, but achieved this speed by demanding significant, temporary surges in the active agent count. Conversely, the decentralized DADENCE architecture demonstrated excellent autonomous resilience. Without requiring any additional algorithmic complexity, the decentralized swarms naturally compartmentalized and healed network fractures efficiently, at the cost of prolonged recovery times. Ultimately, the selection between a centralized or decentralized deployment architecture in dynamically unstable environments must be dictated by the specific hardware budget and time-critical constraints of the mission.

\bibliographystyle{IEEEtran}
\bibliography{references}

\end{document}